\documentclass[letterpaper, 10 pt, conference]{ieeeconf}  

\IEEEoverridecommandlockouts                              

\title{\LARGE \bf Physics-Aware Robotic Palletization with Online Masking Inference}

\author{Tianqi Zhang, Zheng Wu, Yuxin Chen, Yixiao Wang, Boyuan Liang, \\Scott Moura, Masayoshi Tomizuka, Mingyu Ding$^\dag$, Wei Zhan
\thanks{$^\dag$Corresponding author. All authors are with UC Berkeley and Mingyu is also with UNC-Chapel Hill. This work was supported by Berkeley DeepDrive. The authors thank Anyware Robotics for their hardware support and Tim Kehl from ZF for the constructive feedback.}
}
\usepackage{graphicx} 
\usepackage{amsmath}
\usepackage{amssymb}
\usepackage{bbm}
\usepackage{colortbl}
\usepackage{color,xcolor}
\usepackage{booktabs}
\usepackage{subcaption}
\usepackage{times} 

\usepackage{bbold}

\usepackage[para,online,flushleft]{threeparttable}
\usepackage[colorlinks, linkcolor=red]{hyperref}

\definecolor{MyCyan}{RGB}{0,163,218}
\definecolor{MyDarkBlue}{RGB}{0,103,165}
\definecolor{MyDarkGreen}{RGB}{56,116,51}
\definecolor{MyMagenta}{RGB}{200,18,126}

\begin{document}

\maketitle
\thispagestyle{empty}
\pagestyle{empty}


\begin{abstract}
The efficient planning of stacking boxes, especially in the online setting where the sequence of item arrivals is unpredictable, remains a critical challenge in modern warehouse and logistics management.
Existing solutions often address box size variations, but overlook their intrinsic and physical properties, such as density and rigidity, which are crucial for real-world applications.
We use reinforcement learning (RL) to solve this problem by employing action space masking to direct the RL policy toward valid actions.
Unlike previous methods that rely on heuristic stability assessments which are difficult to assess in physical scenarios, our framework utilizes online learning to dynamically train the action space mask, eliminating the need for manual heuristic design.
Extensive experiments demonstrate that our proposed method outperforms existing state-of-the-arts. Furthermore, we deploy our learned task planner in a real-world robotic palletizer, validating its practical applicability in operational settings. The code is available at \url{https://github.com/tianqi-zh/palletization}.
\end{abstract}

\section{INTRODUCTION}
\label{section:introduction}

In modern warehouse and logistics management, stacking boxes continues to be a common challenge. In the past, due to the smaller scale of trade and lower efficiency requirements, workers could rely on their experience to decide how each box should be placed. However, with the globalization of trade, there is a growing need for fast and stable box stacking, and a good solution for this is robotic palletization~\cite{lamon2020towards} \cite{szczepanski2022optimal}.

The task planning for robotic palletization in most industrial environments can be conceptually framed as a variant of the online 3D Bin Packing Problems (BPP), where the inventory of items is predetermined, yet their sequence of arrival remains unpredictable \cite{wang2020robot}. 
Existing works often focus only on boxes' size variations \cite{zhao2021online} \cite{zhao2021learning}, while they overlook the intrinsic differences between boxes. However, in real logistics scenarios, the overall density and rigidity of boxes can vary significantly. Placing a heavy box on top of a soft and lightweight box can pose substantial risks. Therefore, we considered the density and rigidity of the boxes to better align with the demands of real-world logistics scenarios.
Beyond that, we follow \cite{wu2024efficient} and consider a buffer size, which allows the robot to utilize an auxiliary space for storing up to $N$ pending items, thereby expanding its operational capabilities beyond merely handling the immediate one.

In early studies addressing online 3D bin packing problems, hand-coded heuristic methods were primarily used \cite{ha2017online} \cite{wang2019stable}. With the increasing amount of research related to reinforcement learning (RL) \cite{kaelbling1996reinforcement}, more and more studies are exploring how to use RL to solve this problem \cite{zhao2021online} \cite{zhao2021learning}  \cite{wu2024efficient}.

Due to the combinatorial nature of the action space of the problem, applying RL to solve online 3D BPP suffers from the problem of large action space, which will complicate the RL training process \cite{dulac2015deep}. One commonly used solution to the problem of RL with large action space is through ``invalid action masking'' \cite{vinyals2017starcraft} \cite{ye2020mastering}, which identifies and masks out the invalid actions and directs the policy to exclusively sample valid actions during the learning phase.

Our work also adopts action space masking, pinpointing valid actions as task plans that guarantee the stability of intermediate item stacks on the pallet under gravitational forces throughout the training process.
However, the space masking methods in previous studies are not applicable in the setting of this paper.
The heuristic-based methods \cite{zhao2022learning} \cite{faroe2003guided} used in previous work to assess stability become ineffective, as they did not take into account the rigidity and density of the boxes.
Although Wu et al. \cite{wu2024efficient} has employed neural networks to evaluate stability, they still require an initial round of data collection using heuristic methods before training the neural network to serve as action masking model.

To address the reliance on heuristic methods, we propose a new framework that uses online learning to train an action space mask, which entirely eliminates the need for manually designing a heuristic method. Specifically, leveraging the image-like characteristics of observation and action mask data, we employ a semantic segmentation paradigm \cite{chen2014semantic}, \cite{long2015fully} to train the action masking model. The training dataset is compiled through an online data collection phase, utilizing a physics engine to verify the stability of specific placements across various pallet configurations. Moreover, we design the method for collecting and managing online data, allowing the action masking model to efficiently learn the underlying physical principles of box stacking within the RL training loop.

To evaluate the effectiveness of our proposed methodology, we performed a series of comprehensive experiments and compared our results with current RL-based online 3D bin packing solutions. The experimental outcomes reveal that our method significantly surpasses existing baselines. Additionally, we deployed our learned task planner in a real-world robotic palletizer, showcasing the practical applicability of our approach in actual operational environments.

\begin{figure*}[h]  
    \centering
    \includegraphics[width=0.96\textwidth]{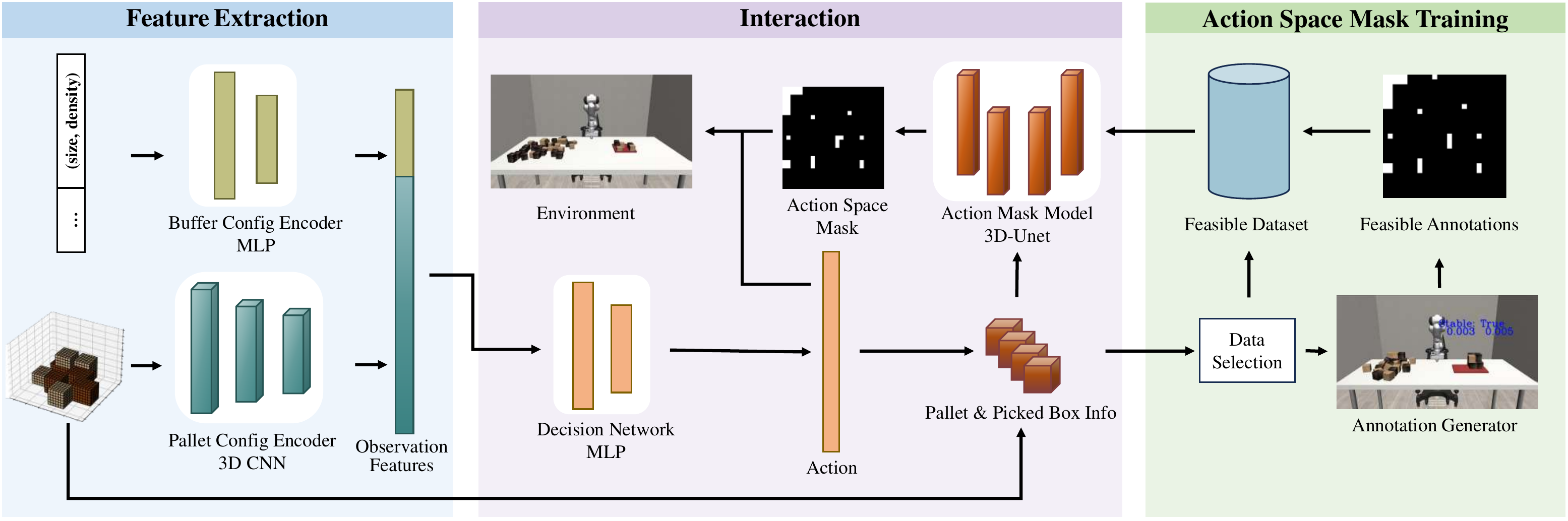}
    \caption{The whole framework of our proposed method. The entire framework can be divided into three parts. The first part is feature extraction. At each timestep t, given the state $s_t = \{C_t, d_t\}$, we process the pallet configuration $C_t$ using a 3D CNN and the properties of the boxes $d_t$ in the buffer using an MLP. These two components are then concatenated to form the observation feature. The second part is interaction. The latter half of the policy network outputs an action based on the observation feature. We use a 3D-UNet \cite{cciccek20163d} as the action masking model. However, we have added a convolutional layer at the end to transform the multi-channel 3D array input into a single-channel 2D map. It generates an action space mask that filters out unstable box placement points, by using the selected box’s properties and the current pallet configuration. The action is then executed in the environment, and the result is obtained. The third part is the training of the action masking model. Unlike the first two parts, which are executed at each timestep, this part only occurs when the policy is updated. First, multiple parallel simulators are used to generate the corresponding feasible annotations for the data selected during the rollout. These data points and their annotations are then appended to the feasible dataset, which is subsequently used to train the action masking model over several epochs.}
    \vspace{-6pt}
    \label{fig:Framework}
    \vspace{-14pt}
\end{figure*}

\section{Related works}
\label{section:related works}

\textbf{Offline 3D bin packing problem. }It is a well-known combinatorial optimization problem in which the objective is to efficiently pack a set of three-dimensional items into one or more bins (or containers) of fixed dimensions, maximizing space utilization. In the offline version, all the items and their dimensions are known in advance before packing begins, which has been particularly challenging due to its NP-hard nature. For smaller instances, methods such as integer linear programming (ILP) or branch-and-bound can be applied to find optimal solutions. To deal with a larger number of boxes, early research focuses on heuristic and meta-heuristic methods \cite{faroe2003guided} \cite{crainic2009ts2pack} \cite{kang2012hybrid}. More recently, deep reinforcement learning has been applied to tackle this task \cite{hu2020tap} \cite{zhang2021attend2pack}.

\textbf{Online 3D bin packing problem. }This is a variation of the 3D BPP where items arrive sequentially over time, requiring placement decisions to be made without knowledge of future items. Unlike the offline version, where all items are known beforehand, this uncertainty necessitates a balance between immediate efficiency and maintaining flexibility for future items, and algorithms must be adaptable to varying patterns and sizes of incoming items. To address this problem, researchers also begin with adaptive heuristics algorithms in the early stage. Initially, solutions are based on deep-bottom-left (DBL) heuristics\cite{karabulut2004hybrid} \cite{ha2017online}, later evolving to the Heightmap-Minimization technique \cite{wang2019stable}. Nowadays, deep reinforcement learning (DRL) has become popular for addressing this problem, though it is challenged by large action space. Zhao et al. \cite{zhao2021online} address this challenge by implementing straightforward heuristics to reduce the action space, in another work they \cite{zhao2022learning} seek to alleviate this issue by devising a packing configuration tree that employs more intricate heuristics to identify a subset of feasible actions. Wu et al. \cite{wu2024efficient} first attempt to use neural networks as an action space mask, but it also relies on heuristic methods for data collection. 

In this work, we study a variant of the online 3D bin packing problem (BPP). We take the intrinsic properties of the boxes into account, such as density and rigidity. With this consideration, even if a box is fully supported, it will collapse if the box beneath it lacks the rigidity to bear the weight of the box above. As a result, there are fewer stable placement positions for the boxes in our problem compared to the standard online 3D BPP.

\textbf{Action masking model.}
One of the major challenges in using RL to solve the online 3D BPP is the large action space. To address this, action space masking is typically employed to reduce the size of the action space. Previous work \cite{zhao2021online} used heuristic methods as action masking models, often manually adding constraints related to the box's support area and the number of supporting edges, only considering placements that satisfy these constraints as stable. \cite{wu2024efficient} trains a neural network as an action masking model through supervised learning. However, to solve the out-of-distribution problem common in supervised learning, the method in \cite{wu2024efficient} required multiple rounds of training, followed by offline training of the neural network. The first round of RL training still depended on heuristic methods. Therefore, this neural network approach is an improvement over heuristic methods, but it does not eliminate the reliance on them.
\section{Method}
\label{section:Method}
\subsection{Problem Formulation}
The objective of palletization is to maximize space utilization while ensuring stability. Space utilization refers to the proportion of the total volume occupied by boxes on the pallet relative to the total available space on the pallet. Besides, we define a stable state on the pallet as follows: when the agent places a box, we record the spatial position and orientation of the box at the moment of placement. At each timestep, after placing a new box, we check the difference between the current position and orientation of all boxes on the pallet and their recorded placement positions. If the deviation for any box exceeds a certain threshold, the state is considered unstable.

\textbf{State.} The state $s_t = \{C_t,d_t\}$ is composed of two parts: the first part is the pallet configuration $C_t$, and the second part is the properties of the boxes in the buffer $d_t$.

We represent the pallet's state using voxels, since we need to account for the intrinsic properties of the boxes. More specifically, in addition to the dimensions of the boxes, we also consider their density and rigidity. Therefore, we quantify the pallet configuration with a 2-channel 3D array, stands for the density distribution and rigidity distribution. $d_t$ is a 1D array composed of information from $N$ boxes, where each box's properties include length, width, height, density, and rigidity. Therefore, it is an array of size $5N$, where $N$ represents the buffer capacity.

\textbf{Action.} It should include the following information: which box from the buffer to select, how to rotate the box, and where to place it on the pallet. We assume that after rotation, each box remains parallel to the boundary faces of the pallet, resulting in six possible orientations for each box. Additionally, we consider that new boxes cannot be placed unsupported in mid-air, so the placement is determined by the $(x,y)$ coordinates. We discretize the pallet's length and width into $l_p$ and $w_p$, respectively, giving $l_p \times w_p$ possible placement positions for each box.

As a result, the action $a$ belongs to $ \mathbb{R}^{N \times 6 \times l_p \times w_p} $. For instance, with $N=5$ and $l_p=w_p=25$ \cite{wu2024efficient}, this results in 18,750 possible outcomes for the action. The vast size of this action space greatly increases the complexity of the RL problem, making the optimization process substantially more challenging.

\subsection{Reward Design}
We encourage the planner to improve space utilization, as long as it is stable. Therefore, we use the proportion of the box's volume placed at each step relative to the total space as the reward for each step. The reward function we designed is as follows:
\begin{equation}
    r(s_t) = \mathbb{1}(s_t) \cdot \frac{V_{box}}{V_{max}}
\end{equation}
$V_{box}$ refers to the volume of the box selected at timestep $s_t$, while $V_{max}$ represents the theoretical maximum volume that the pallet can accommodate. $\mathbb{1}(s_t)$ serves as an indicator function that validates the physical stability of the boxes on the pallet at state $s_t$.

In addition, to improve efficiency, we introduced a penalty term. When the number of boxes in the buffer is less than $N$, if the selected index does not correspond to any box, then $r(s_t)= -1$.

Another factor that affects the overall reward is the episode length. There are three reasons for an episode to end: after selecting a box, the action masking model determines that no feasible placement points exist; placing the box results in instability; or all the boxes have been placed on the pallet.

\subsection{Integrating the Action Masking Model into the RL training process}
At each timestep $t$, after the policy outputs an action, the corresponding box is selected from the buffer based on the action's information. The box's properties are then combined with the current pallet configuration to form a multi-channel 3D array, which is fed into the action masking model. The action masking model outputs an action space mask, indicating which points are considered stable and which are unstable at that moment. We refer to the set of stable points as the `feasible set'. The model finds the point closest to the action’s chosen position within the feasible set (if there are multiple nearest points, one is randomly selected). The box is then placed in the environment at this selected point within the feasible set.

\subsection{Online action masking model learning}
We choose to use a neural network as the action masking model and train it using an online approach because heuristic methods have numerous limitations that render them inadequate for our task setting.

Heuristic methods inherently face many limitations when addressing normal online 3D BPP. For example, the necessity for multiple hyperparameters in most action masking methods poses a challenge in tuning these parameters to accommodate diverse scenarios, and they fail to account for uncertainties inherent in real-world palletization execution \cite{wu2024efficient}.
When considering the density and rigidity of the boxes, the heuristic methods used in previous works become unusable. In the original scenario, the feasible set identified by the heuristic action masking model was a subset of the true feasible set. However, in this new scenario, they become an intersection, making the heuristic action masking model highly inefficient. This inefficiency is evident in the experimental results presented in Section \ref{section:experiment}. Designing a new heuristic method that incorporates both density and rigidity manually is cumbersome and lacks generalizability. 

Therefore, we need a new paradigm to solve this problem, one that relies on directly acquiring experience from the simulator. We propose the online action space mask learning paradigm, which incorporates supervised learning for the action masking model into the RL training loop, shown in Figure \ref{fig:Framework}. Initially, both the policy network and the action masking model are randomly initialized. During the RL policy rollout, we record data for action masking model training. Those data contain two parts: the configurations of the pallet $C_t$ and the properties of the next to be placed box $b_t$. 
After a certain number of timesteps, we update the action masking model along with the policy network. The update process for the action masking model involves two steps: first, generating an annotation $g_t$ for each recorded data point $(C_t, b_t)$ from the previous rollout phase; second, appending these data points along with their annotations to the dataset, called feasible dataset, and using this feasible dataset to train the action masking model.

We use multiple parallel simulators specifically created to generate annotations. In these simulators, the scene is first reconstructed based on the assigned $(C_t, b_t)$, after which the box is tested at every possible position. It is important to note that some positions can be judged using simple heuristics without requiring the simulator, such as a box placed directly on the pallet surface being inherently stable, or a box with less than 25\% support being inherently unstable.

There are some important details regarding data selection and management. For episodes that end due to instability, the corresponding data is always recorded. For regular rollout data, there is a 0.1 probability of being recorded. This ensures that the action masking model focuses on learning cases where its predictions were less accurate, while still capturing regular data to avoid overfitting. For managing the feasible dataset, we use a deque data structure, meaning that data is managed in a first-in-first-out manner. This allows the action masking model to encounter more diverse data, helping to address the issue of out-of-distribution data.

We demonstrate the effectiveness of our proposed method in the following section. 
\section{EXPERIMENTAL VALIDATIONS}
\label{section:experiment}

\begin{figure}[t]  
    \centering
    \includegraphics[width=0.7\linewidth]{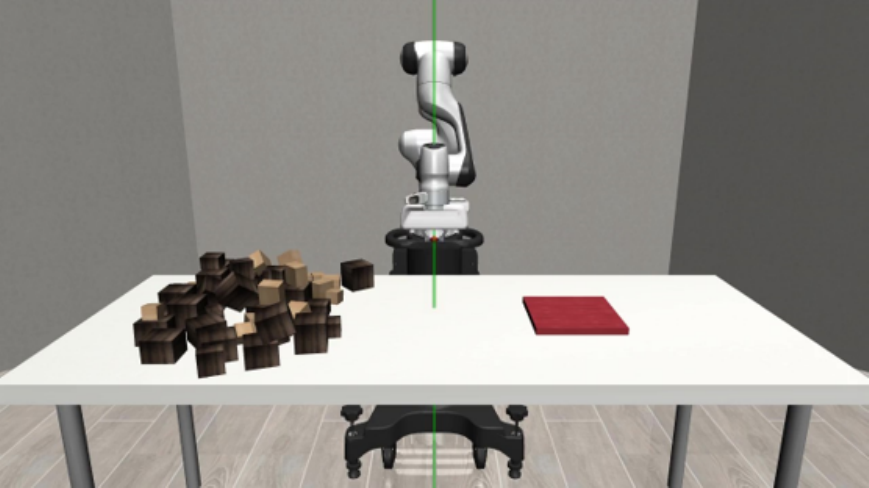}
    \caption{Visualization of our simulated palletization environment in MuJoCo \cite{todorov2012mujoco}. Although 40 boxes are displayed for illustrative purposes, the robot is programmed to perceive and interact with only N boxes within the buffer area. The arrangement of the boxes is randomized and unknown, shuffled anew for each RL episode.}
    \vspace{-10pt}
    \label{fig:task_setting}
\end{figure}

In this section, we present a thorough quantitative analysis of our proposed approach, and formulate our experiments to answer the following questions:
\begin{enumerate}
    \item Does the action space mask training method mentioned earlier eventually converge and learn a relatively accurate result?
    \item How are the changes in the performance of the policy and the action masking model related?
    \item Does the online action masking model learning method enhance the RL training process, resulting in a superior policy?
\end{enumerate}

\subsection{Simulation Experiments}
\textbf{Setup.} To explore the task planning challenges associated with the palletization problem, we developed a simulated palletization task in MuJoCo \cite{todorov2012mujoco} \cite{robosuite2020}, reflecting a realistic logistic scenario, as depicted in Figure \ref{fig:task_setting}. We adjusted the parameters in the simulation environment that solve for the box deformation under contact forces to give the boxes different levels of rigidity. There are four different types of boxes in the simulation environment: two soft, low-density boxes and two hard, high-density boxes. The size, density, rigidity, and quantity of boxes in the simulation are shown in Table \ref{tab:box_specifications}.

If a hard box is placed on top of a soft box in the simulation environment, the latter will collapse, which is considered an unstable situation. The pallet is specified to have dimensions of 25 $\times$ 25 inches, with a stipulation that the maximum height of stacked boxes cannot exceed 20 inches. For our experiments, we applied a discretization resolution of 1 inch to both the boxes and the pallet. In all the subsequent simulation experiments mentioned, we have chosen a buffer size of $N=5$. And we use PPO \cite{schulman2017proximal} as the RL algorithm for all experiments.

\begin{table}[t]
    \centering
    \caption{Box Specifications in the Simulation Environment.}
    \label{tab:box_specifications}
    \begin{tabular}{cccc}
        \toprule
        Dimension (inches) & Density & Rigidity & Counts\\
        \midrule
        6 $\times$ 6 $\times$ 4 & 500 & 0.5 & 10\\
        6 $\times$ 6 $\times$ 6 & 500 & 0.5 & 10\\
        6 $\times$ 6 $\times$ 6 & 5000 & 3 & 10\\
        8 $\times$ 6 $\times$ 6 & 5000 & 3 & 10\\
        \bottomrule
    \end{tabular}
    \vspace{-9pt}
\end{table}

At each step of the planning process, the task planner receives the current pallet configuration, represented as a density distribution, along with the properties of the next N boxes in the buffer, which include their dimensions and density. The planner’s task is to select one of the N boxes, determine its orientation, and place it on the pallet. Since our focus is on planning rather than manipulation, the actual `pick-rotate-place' actions by the robot are omitted. Instead, the selected box is immediately positioned at the goal pose determined by the planner, to accelerate the learning process.

To better simulate real-world palletization, we added noise to the box placement when generating feasible annotations: positional noise on the xy plane, $\delta_t \sim N(0, 0.05)$, measured in inches, and rotational noise around the z-axis, $\delta_r \sim N(0,5)$, measured in degrees.

\begin{figure*}[t]
\begin{minipage}[t]{0.32\linewidth}
    \centering
    \includegraphics[width=1\textwidth]{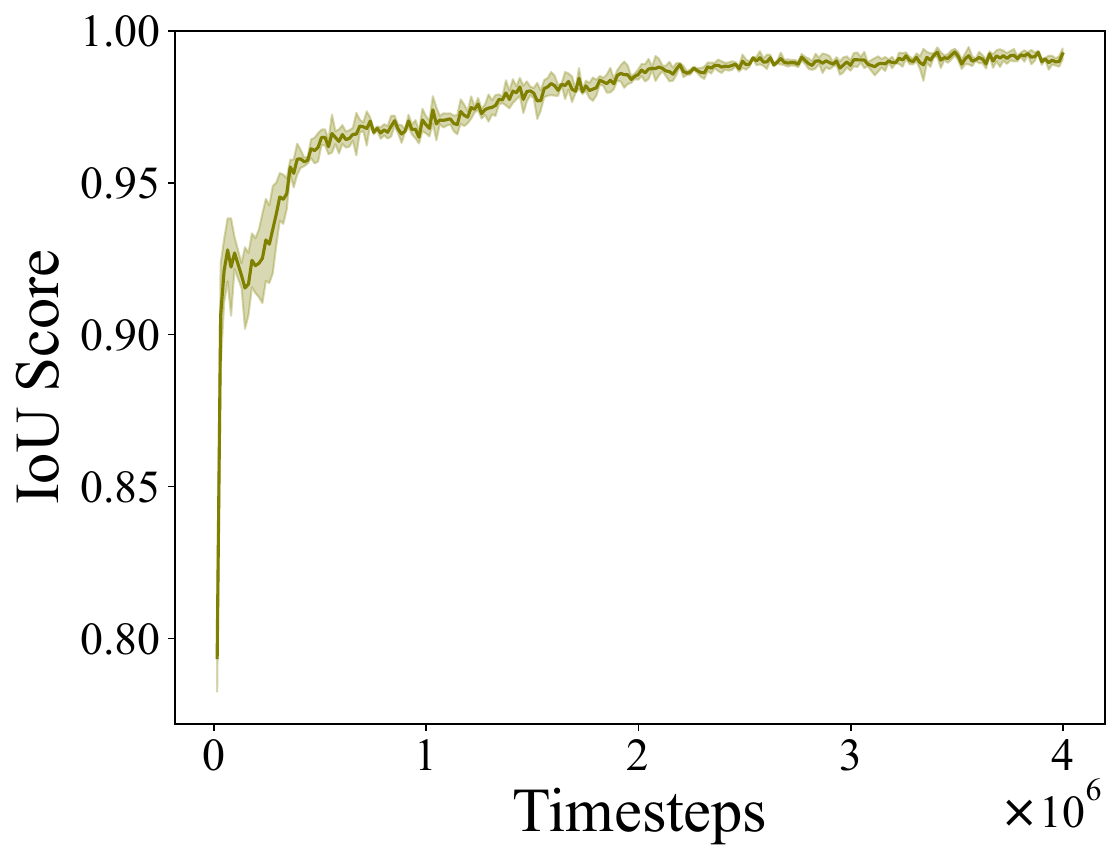}
    \vspace{-16pt}
    \caption{IoU score on validation set. After rolling out for a certain number of timesteps, we update the action masking model and calculate the IoU score on the validation set. Results are averaged over 5 random seeds.}
    \label{fig:val_iou}
\end{minipage}
\hfill
\begin{minipage}[t]{0.335\linewidth}
    \centering
    \includegraphics[width=1\textwidth]{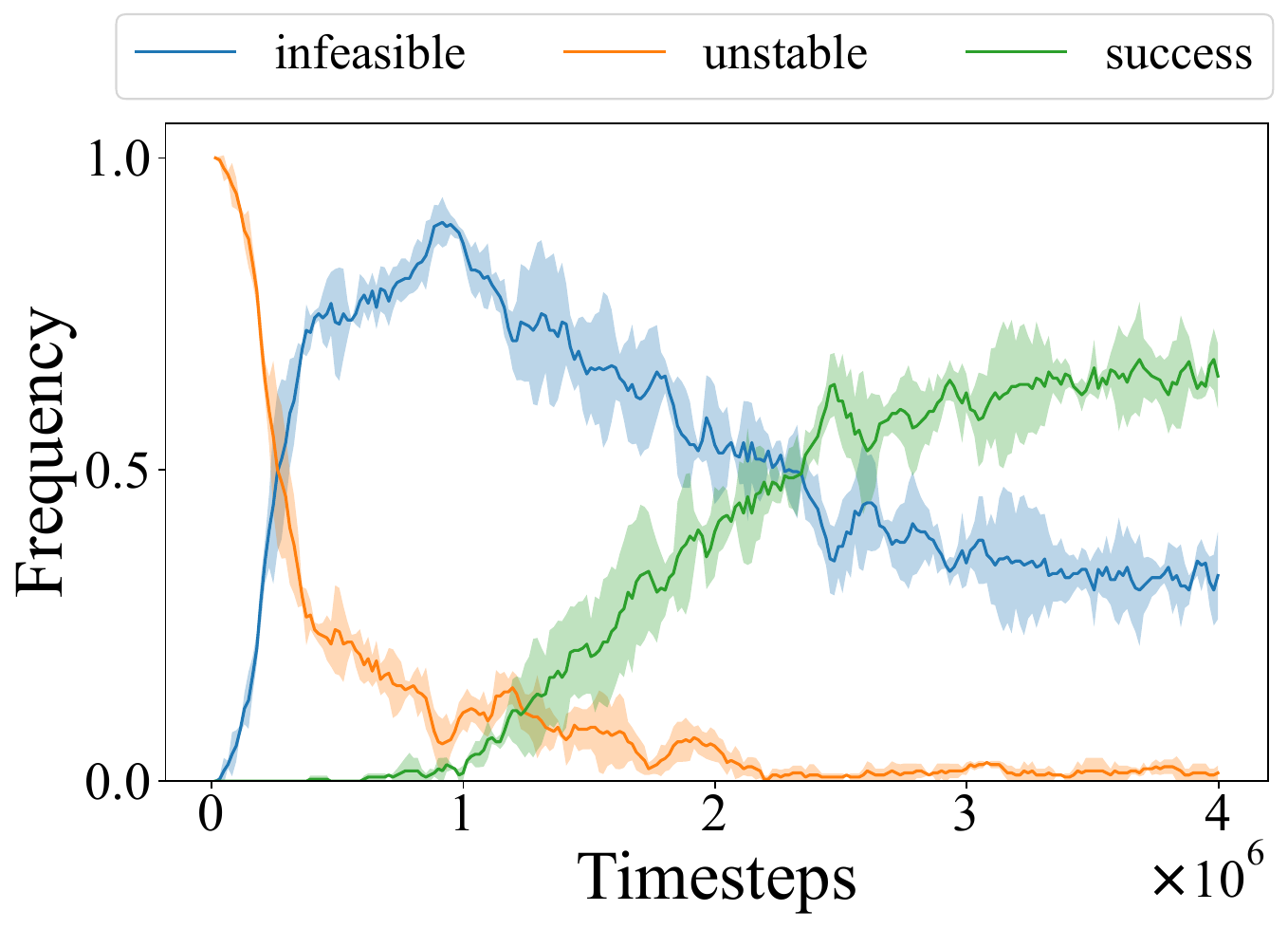}
    \vspace{-16pt}
    \caption{The frequency change curve for the three types of episode end reasons. During training, we record the frequency of each episode end reason, which is represented by the curve shown. Results are averaged over 5 random seeds.}
    \label{fig:end_reason}
\end{minipage}
\hfill
\begin{minipage}[t]{0.32\linewidth}
    \centering
    \includegraphics[width=1\textwidth]{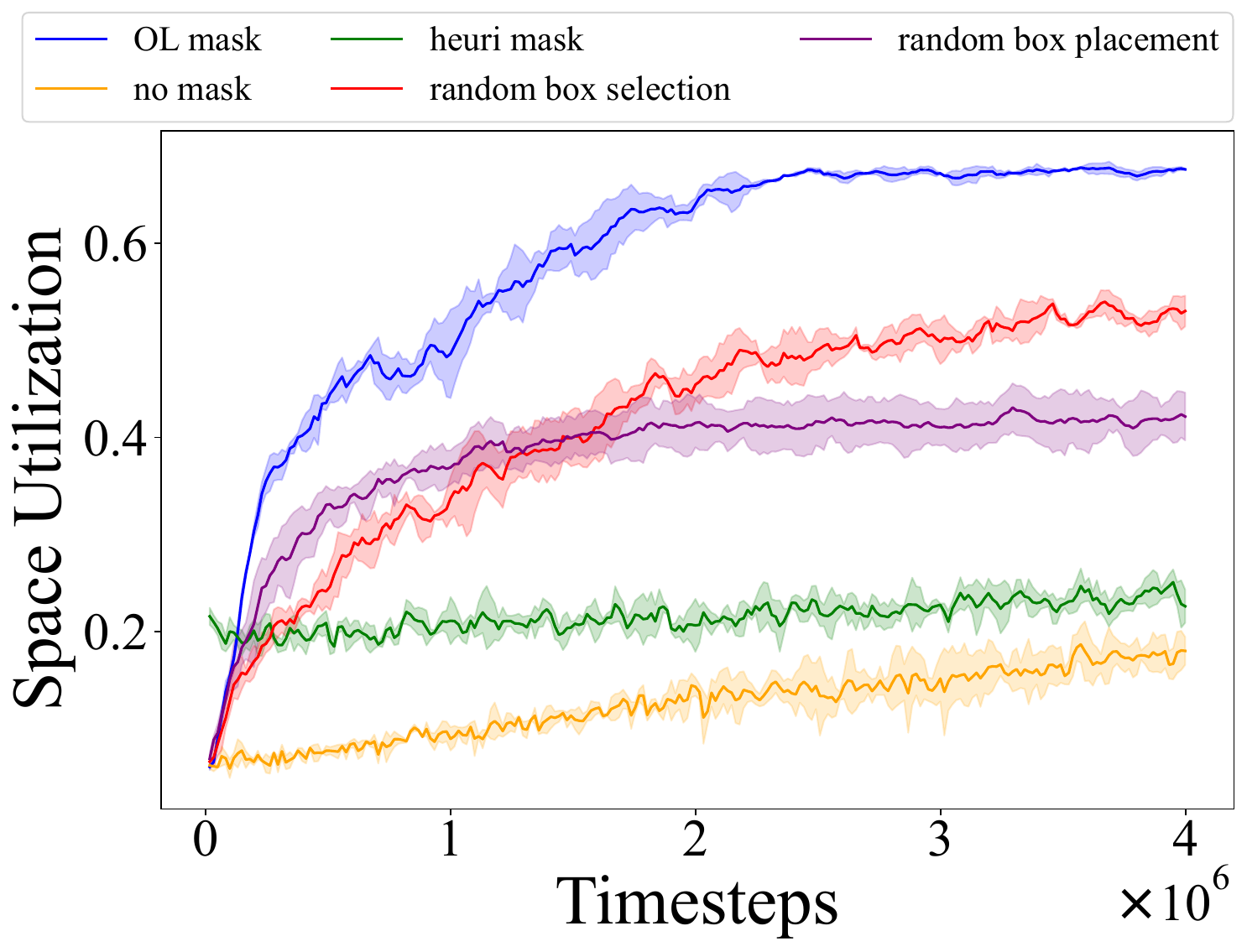}
    \vspace{-16pt}
    \caption{Space utilization. Our method (OL Mask) demonstrates faster convergence and achieves better space utilization compared to the other methods. Results are averaged over 5 random seeds.}
    \label{fig:space_utilization}
\end{minipage}
\end{figure*}

\begin{table*}[t]
    \centering
    \caption{Performance evaluation of different planners under various methods.}
    \label{tab:final performance}
    \setlength{\tabcolsep}{12pt}
    \resizebox{\linewidth}{!}{
    \begin{tabular}{lccccc}
        \toprule
        ~ &  Nomask & Random Selection & Random Placement & Heuristic mask & \textbf{OL Mask (Ours)} \\
        \midrule
        \textbf{Space utilization}  & 0.18 & 0.53 & 0.41 & 0.26 & \textbf{0.68} \\
        \textbf{Infeasible Rate} & \textbf{0.00} & 0.84 & 0.95 & 0.05 & 0.34 \\
        \textbf{Unstable Rate} & 1.00 & 0.14 & 0.05 & 0.95 & \textbf{0.01} \\
        \textbf{Mean Episode Length} & 12.3 & 31.8 & 25.8 & 15.9 & \textbf{39.4} \\
        \textbf{Mean Episode Reward} & 0.18  & 0.49 & 0.41 & 0.26 & \textbf{0.66} \\
        \textbf{Success Rate} & 0.00 & 0.02 & 0.00 & 0.00 &  \textbf{0.65} \\
        \bottomrule
    \end{tabular}
    }
    \vspace{-6pt}
\end{table*}

\textbf{Action space mask training results. }
During the training of the policy, we used a deque with a size of 16,000 as a feasible dataset for training the action masking model.

For each policy rollout step, if the step is unstable and causes the episode to end, that data will definitely be recorded in the dataset. If it is just a normal step, there is a 10\% chance it will be recorded in the dataset. In this way, we can make the action masking model focus more on cases where it failed to make correct judgments, ultimately leading to more accurate performance.

We split the dataset into an 8:2 ratio for the train set and validation set, using the model's IoU score on the validation set to evaluate its performance.
At each policy update, we simultaneously update the action masking model. 
The IoU score is very low at the beginning. As the total number of training epochs increases and the model encounters more new data, the capability of the action masking model gradually improves. The action masking model using the online learning method began to converge after approximately 2 million rollout timesteps, as shown in Figure \ref{fig:val_iou}. Eventually, the IoU score converges to around 98\% on the validation set

Such a high IoU score indicates that the action masking model is already capable of determining which positions are stable or unstable based on the current situation of boxes on the pallet and the information of the next box.



\textbf{Guidance from the action space mask. } To study the relative performance of the action masking model and the policy, we recorded the probability of each of the three end reasons occurring in each episode during training, as shown in Figure \ref{fig:end_reason}. Each episode can be ended for the following reasons. 
\begin{itemize}
    \item \textit{Infeasible:} The action masking model indicates that there is no feasible point on the pallet; \item \textit{Unstable:} After placing each box, the situation on the pallet becomes unstable, resulting in a collapse.
    \item \textit{Success:} All 40 boxes have been placed properly.
\end{itemize}
Initially, because the action masking model is randomly initialized, its judgments on stability are very inaccurate. As a result, in the beginning, every episode ends as unstable. However, as the action masking model continues to be trained, its accuracy gradually improves. At this point, since the policy has not yet learned how to place boxes densely, the frequency of cases where boxes are selected but have no place to be placed increases significantly. Finally, as the policy, guided by the accurate action space mask, continuously attempts different placements, it learns a compact and stable way to place boxes, leading to a significant increase in the success rate. Since the training of the action masking model is supervised learning, it is more likely to achieve better performance more quickly. Therefore, it first converges and then guides the policy to achieve good performance.


\begin{figure*}[h]  
    \centering
    \includegraphics[width=0.93\textwidth]{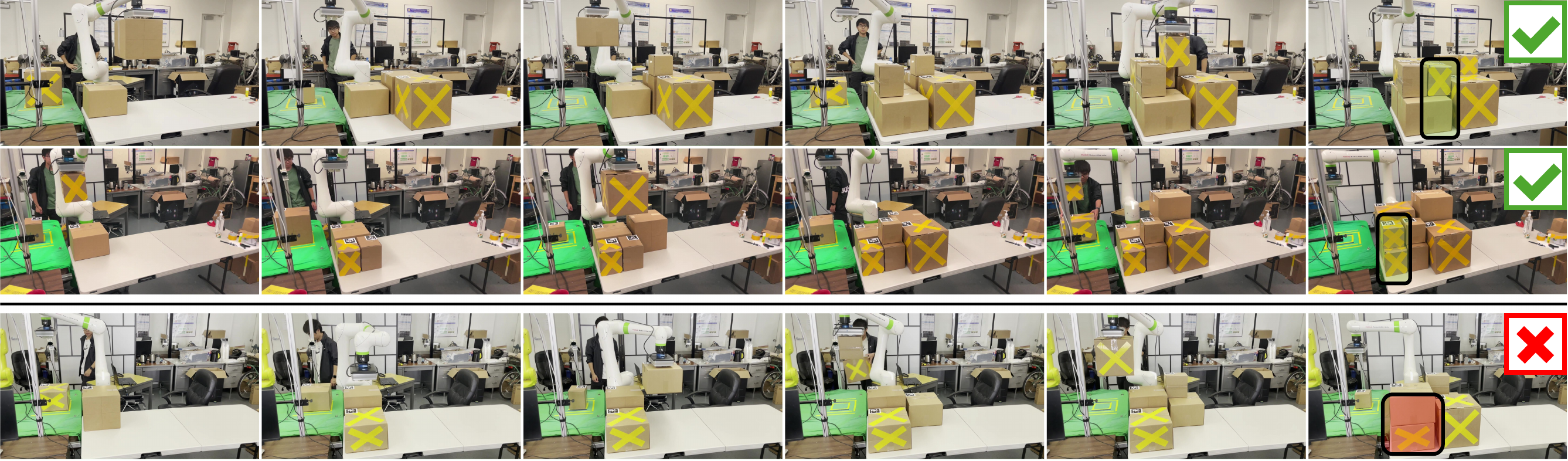}
    \vspace{-4pt}
    \caption{Real-world experiments with each row as a planning sequence. The boxes with yellow cross tape are considered to have low density and be soft, while other boxes are considered to have high density and be hard.
    1) The first two rows show the planner trained using our method, successfully placing 9 boxes on the pallet while maintaining stability, under two different box arrival sequences. 
    2) The bottom row shows the planner trained with a heuristic mask, which, after stacking 5 boxes, incorrectly placed a high-density, high-rigidity box on top of a low-density, low-rigidity box, which is considered a failure case of unstable. The unstable position is marked with a red cross.}
    \label{fig:frame_sequence}
    \vspace{-12pt}
\end{figure*}

\textbf{Ablative experiments. }
To verify the importance of both RL planning and online action space mask learning in our method, we selected four baselines.
\begin{itemize}
  \item Nomask: Action masking is not applied during the RL training process.
  \item Heuristic mask: We use the heuristic mask from \cite{zhao2021online}, which considers two criteria: the support ratio and the number of support corners. A placement is considered stable in any of the following three cases: $\text{support ratio}>0.6 \land \text{support corners} = 4$; $\text{support ratio}>0.8 \land \text{support corners}\geq 3$; $\text{support ratio}>0.95$
  \item Random box selection: In this method, an online training action space mask is used, but the selection of boxes is independent of the actions output by the RL algorithm and is instead randomly drawn from the buffer.
  \item Random box placement: Under the premise of using an online training action space mask, the selection of boxes is determined by the policy, but their placement is random.
\end{itemize}

We use the mean episode space utilization to measure the overall quality of the final policy, as shown in Figure \ref{fig:space_utilization}. 
Initially, the space utilization of the three methods using online action masking model learning (`OL mask', `random box selection', `random box placement') is the same as that of the no mask method, due to their action masking models being randomly initialized. At the beginning, the heuristic mask method achieves the highest space utilization. However, as training progresses, the space utilization of the three online action masking methods quickly surpasses that of the other two methods.
Among all the methods, the space utilization for the `nomask' method remains the lowest throughout the training process, while the space utilization for the `nomask' method remains the highest. After 4M timesteps, the three methods using the online learning action space mask perform significantly better than the other two methods. This demonstrates the importance of the action space mask and shows that an accurate action space mask can accelerate convergence and greatly improve the planner's capability. Comparing our method with the `random box' and `random place' methods also highlights the importance of RL in the overall approach, demonstrating that each component of our method is indispensable as part of the whole.

\subsection{Real World Experiments}
We further implement the policy and action masking model learned in simulation onto a real robot arm. The gripper used in our lab environment is relatively large, which requires the boxes to also be larger. Additionally, due to the limited range of motion of the robotic arm, it is not possible to stack 40 boxes as we did in the simulation when working with larger boxes. However, in a real-world logistics warehouse with larger robotic arms, it would be possible to replicate the scenario from the simulation. We also train a new policy and a new action masking model to accommodate the real-world setting.

Since we focus solely on the planning aspect of palletization, we assume that the perception task should be handled earlier in the pipeline. We use AprilTags \cite{wang2016apriltag} to record the size, density, and softness of each box. Near the buffer, we installed an Intel RealSense camera, which scans the AprilTags and compares the data with pre-stored information in the computer to obtain the details of the boxes currently in the buffer. Due to spatial constraints, we set the buffer size to 1. Additionally, since the robotic arm can easily rotate boxes around the z-axis during operation, we only consider rotation around the z-axis.

\begin{figure}[h]
    \centering
    \includegraphics[width=0.3\textwidth]{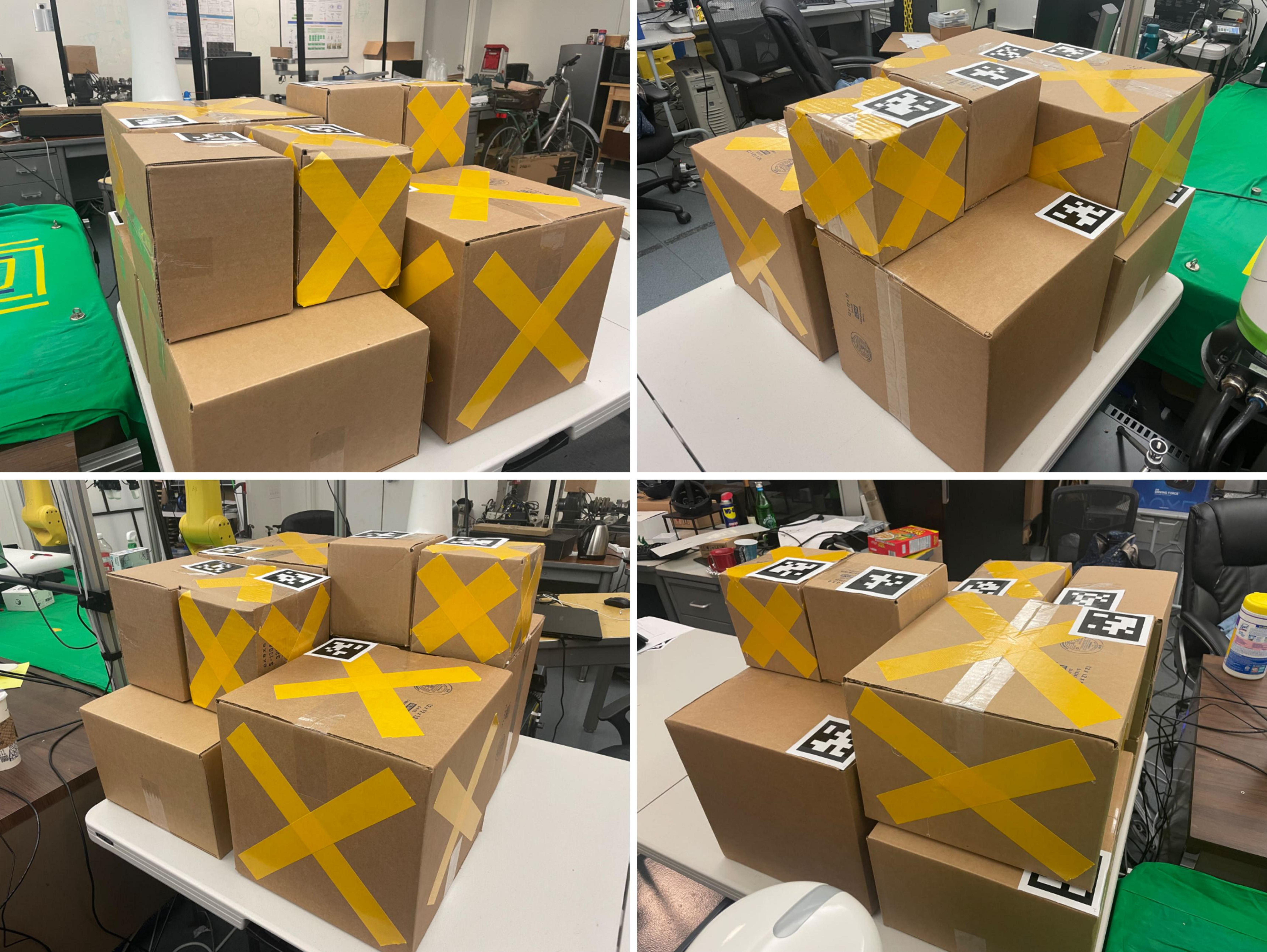}
    \caption{Final pallet configuration viewed from 4 perspectives. In a stable situation, boxes with high density and rigidity should be placed under heavy and soft ones.}
    \vspace{-6pt}
    \label{fig:real_world_4p}
\end{figure}

The pallet we considered has both a length and a width of 24 inches, with a maximum allowable stacking height of 20 inches. We consider a total of 9 boxes, but there are 8 different types of boxes. Finally, our planner successfully places all 9 boxes on the pallet and achieves a stable solution, with a space utilization rate of 70\%.
The compact and stable configuration highlights the effectiveness and reliability of the learned task planner, shown in Figure \ref{fig:real_world_4p}.

\section{Conclusion and future work}
\label{section:conclusion}

In this paper, we introduce a variant of the online 3D bin packing problem that incorporates intrinsic box properties beyond dimensions, to more closely resemble real-world scenarios. To address this new challenge, we propose an online action masking model learning method, which completely eliminates the need for heuristic methods, offering a more general approach. Simulations validate the effectiveness of our method, showing superior performance compared to other approaches. In the future, we may explore even more realistic scenarios, such as handling friction between boxes and assuming the pallet may experience sudden shaking. 

{\small
\bibliographystyle{IEEEtran}
\bibliography{references}
}

\end{document}